\documentclass[sigconf]{acmart}

\usepackage{amsmath,amsfonts}
\usepackage{xcolor,colortbl}
\usepackage{multirow}
\usepackage{pifont}
\usepackage{booktabs}

\definecolor{geolinkblue}{RGB}{34,94,168}
\definecolor{geolinkshade}{RGB}{237,243,252}
\newcommand{\best}[1]{\textbf{#1}}
\newcommand{\second}[1]{\underline{#1}}
\newcommand{\cmark}{\textcolor{geolinkblue}{\ding{51}}}
\newcommand{\xmark}{--}

\AtBeginDocument{%
  }

\copyrightyear{2026}
\acmYear{2026}
\setcopyright{cc}
\setcctype{by-nc-nd}
\acmConference[MM '26] {Proceedings of the 34th ACM International Conference on Multimedia}{November 10--14, 2026}{Rio de Janeiro, Brazil.}
\acmBooktitle{Proceedings of the 34th ACM International Conference on Multimedia (MM '26), November 10--14, 2026, Rio de Janeiro, Brazil}
\acmISBN{979-8-4007-2213-4/2026/11}
\acmDOI{10.1145/3767308.3835330}

\begin{document}

\title{GeoLink: A 3D-aware Framework to Improve Generalization for Cross-view Geo-localization}

\author{Hongyang Zhang}
\authornote{Both authors contributed equally to this research.}
\affiliation{%
  \institution{The Chinese University of Hong Kong, Shenzhen}
  \city{Shenzhen}
  \country{China}}
\email{hongyangzhang1@link.cuhk.edu.cn}

\author{Yinhao Liu}
\authornotemark[1]
\affiliation{%
  \institution{Xiamen University}
  \city{Xiamen}
  \country{China}}
\email{liuyinhao28@stu.xmu.edu.cn}

\author{Haitao Zhang}
\affiliation{%
  \institution{Xiamen University}
  \city{Xiamen}
  \country{China}}
\email{haitaozhang0829@gmail.com}

\author{Zhongyi Wen}
\affiliation{%
  \institution{University of Electronic Science and Technology of China}
  \city{Chengdu}
  \country{China}}
\email{zy\_wen@std.uestc.edu.cn}

\author{Zhenyu Kuang}
\affiliation{%
  \institution{Foshan University}
  \city{Foshan}
  \country{China}}
\email{kmustkzy@126.com}

\author{Shuxian Liang}
\authornote{Corresponding authors.}
\affiliation{%
  \institution{Tongji University}
  \city{Shanghai}
  \country{China}}
\email{sxliang@tongji.edu.cn}

\author{Xian-Sheng Hua}
\authornotemark[2]
\affiliation{%
  \institution{Tongji University}
  \city{Shanghai}
  \country{China}}
\email{xshua@tongji.edu.cn}

\renewcommand{\shortauthors}{Hongyang Zhang et al.}

\begin{abstract}
Generalizable cross-view geo-localization aims to match the same location across views in unseen regions and conditions without GPS supervision. Its core difficulty lies in severe semantic inconsistency caused by viewpoint variation and poor generalization under domain shift. Existing methods mainly rely on 2D correspondence, but they are easily distracted by redundant shared information across views, leading to less transferable representations. To address this, we propose GeoLink, a 3D-aware semantic-consistent framework for Generalizable cross-view geo-localization. Specifically, we offline reconstruct scene point clouds from multi-view drone images using VGGT, providing stable structural priors. Based on these 3D anchors, we improve 2D representation learning in two complementary ways. A Geometric-aware Semantic Refinement module mitigates potentially redundant and view-biased dependencies in 2D features under 3D guidance. In addition, a Unified View Relation Distillation module transfers 3D structural relations to 2D features, improving cross-view alignment while preserving a 2D-only inference pipeline. Extensive experiments on multiple benchmarks show that GeoLink consistently outperforms state-of-the-art methods and achieves superior generalization across unseen domains and diverse weather environments.
\end{abstract}

\ccsdesc[500]{Computing methodologies~Visual content-based indexing and retrieval}
\ccsdesc[500]{Information systems~Top-k retrieval in databases}

\keywords{Generalizable Cross-view Geo-Localization, 3D-aware Representation Learning, Relational Distillation}

\maketitle

\section{Introduction}
\begin{figure}[t!]
    \centering
    \includegraphics[width=0.48\textwidth]{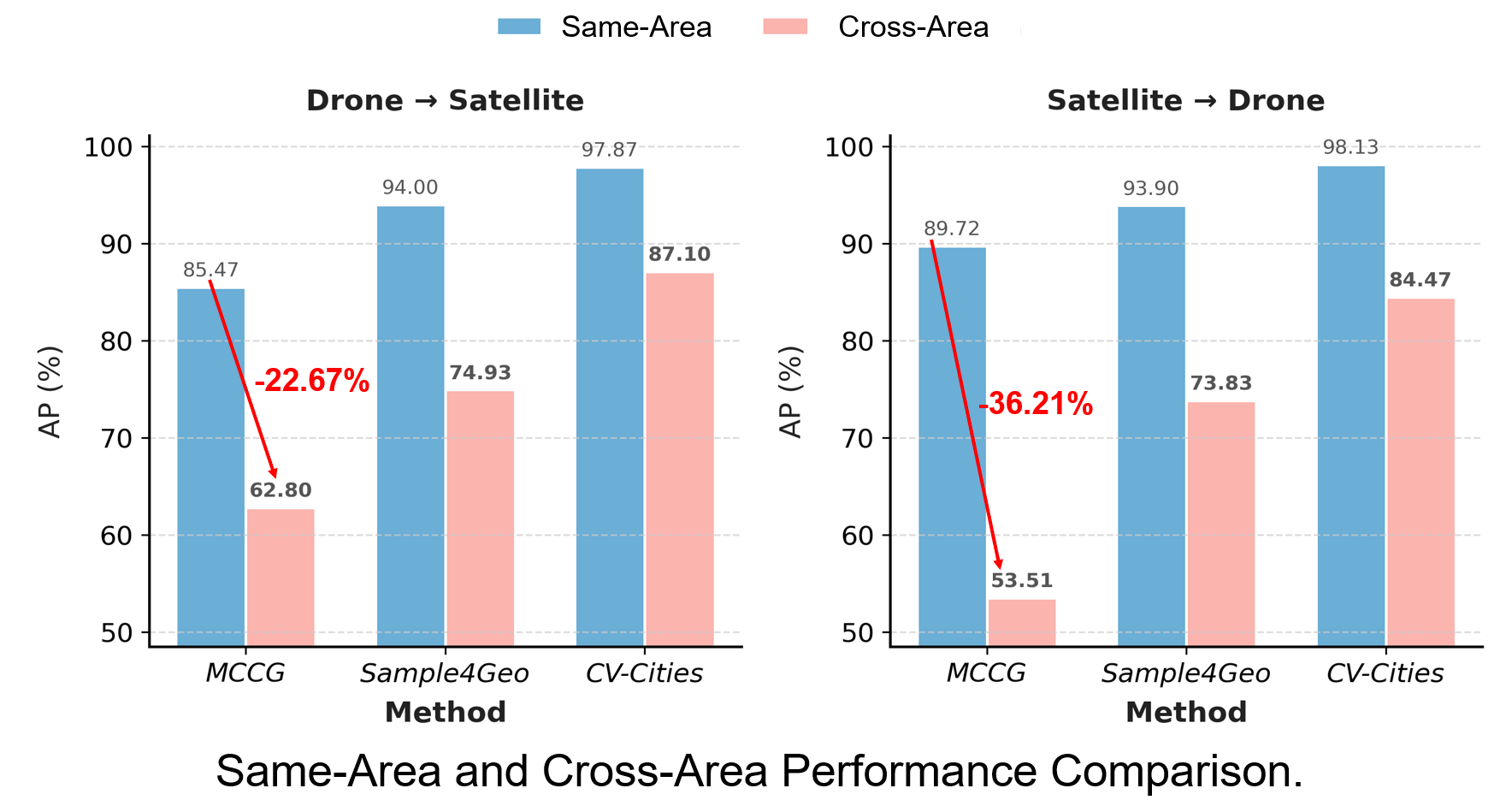}
    \caption{Performance comparisons on SUES-200@(150m). Existing methods perform well in the same-area setting but degrade noticeably in cross-area evaluation, revealing limited generalization ability under domain shift.}
    \label{fig:2}
\Description{Comparison plot showing same-area and cross-area performance for methods on SUES-200.}
    \vspace{-10pt}
\end{figure}

Cross-view geo-localization (CVGL) is formulated as a retrieval task that geolocates multi-view imagery (e.g., drone and satellite) in GPS-denied environments~\cite{r42}, aiming to identify geographically corresponding images from different views using a single-view query. As shown in Figure~\ref{fig:2}, although existing methods~\cite{r5, r69, r78} have achieved promising performance in the same-area setting, they mainly rely on location labels for training. And they often degrade noticeably when deployed in unseen regions due to different scene layouts and building styles, limiting practical applications such as autonomous driving~\cite{r67} and robotic navigation~\cite{r68}. Therefore, CVGL tasks face two core challenges: maintaining \emph{semantic consistency} under drastic viewpoint changes and achieving robust \emph{generalization under domain shift}.

To address the first challenge, existing CVGL methods align cross-view features directly in the 2D feature space. Although such alignment helps reduce the view gap, it may also preserve view-shared yet task-irrelevant content (e.g., broad sky regions, or repetitive background textures). Meanwhile, viewpoint changes introduce geometric biases that can alter the visual semantics of the same scene across views. Although Bird's-eye-view (BEV) projection~\cite{r7} and generative models have emerged as promising solutions for modeling spatial correspondences between ground and aerial viewpoints, these methods still rely predominantly on 2D visual correspondence. Consequently, location-discriminative cues may remain entangled with task-irrelevant variations, thereby limiting fine-grained matching and cross-domain generalization. Similar observations in cross-domain retrieval suggest that suppressing such nuisance factors is important for learning transferable visual representations~\cite{r79}.

\begin{figure}[t!]
	\centering
	\includegraphics[width=3.25in,height=2.15in]{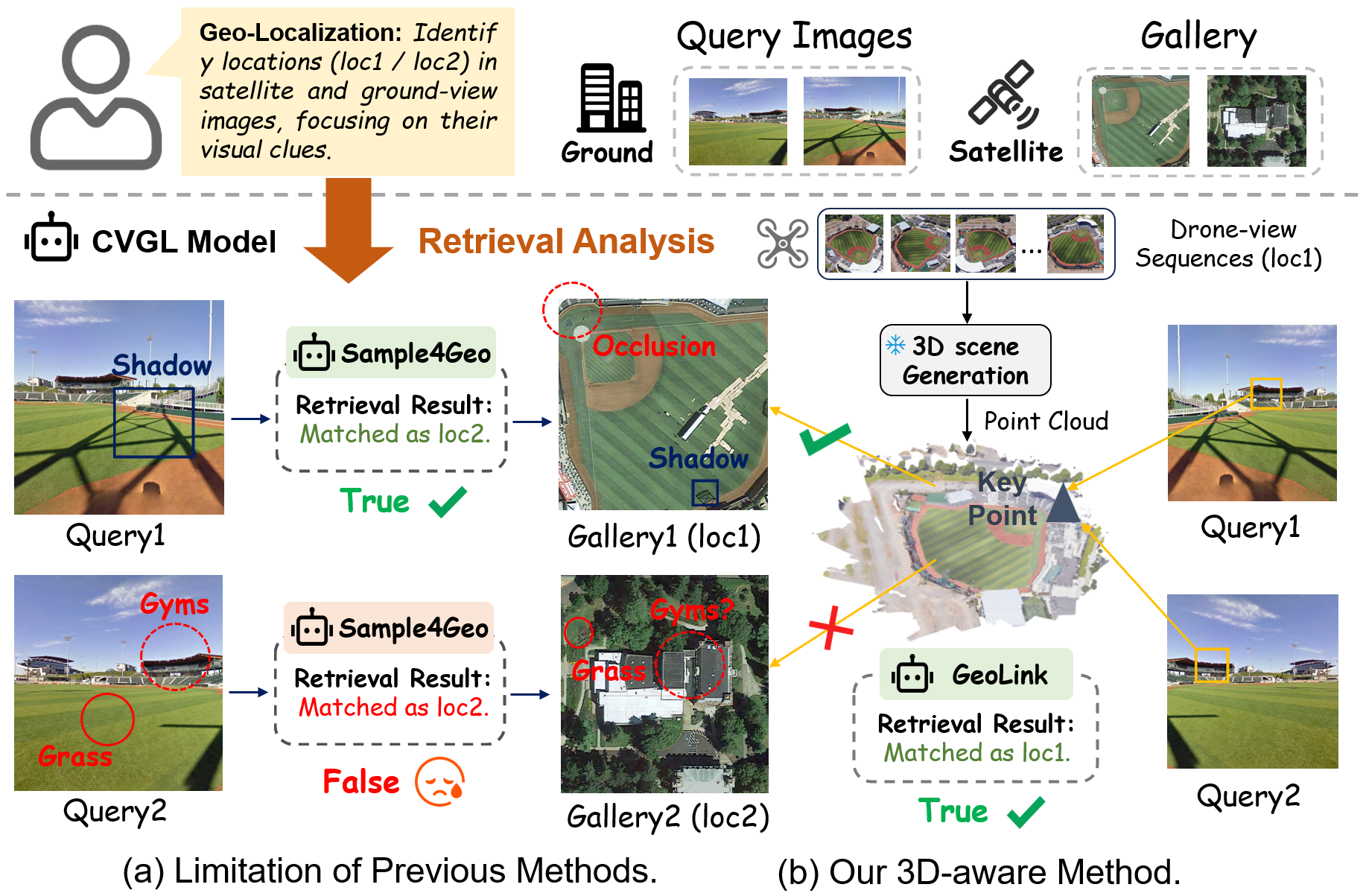}
	\caption{Comparison of previous methods and our approach. (a) Previous methods based only on 2D observations are vulnerable to occlusion, limited FoV, and ambiguous visual cues. (b) Scene-level 3D structure reconstructed from multi-view drone images provides a more stable structural anchor for learning transferable 2D retrieval features.}
	\label{fig:3}
\Description{Motivation illustration comparing view-biased two-dimensional cues with more stable three-dimensional structural cues.}
    \vspace{-10pt}
\end{figure}

To overcome the second limitation, recent studies~\cite{r1, r2} advocate leveraging 3D representations, which provide richer structural and spatial context than 2D cues, enabling more robust cross-view understanding and generalization to unseen scenarios. This observation aligns with human multi-view understanding, which integrates appearance and spatial cues through continuous interaction with 3D environments. As illustrated in Figure~\ref{fig:3}, existing 2D-based methods may over-rely on view-biased appearance cues (e.g., grass), making them vulnerable when visual evidence is ambiguous or partially occluded. In contrast, 3D representations provide more stable and view-consistent structural information. By explicitly modeling scene geometry, they reduce cross-view ambiguity and serve as structural priors for cross-view representation learning.

Despite these advances, achieving robust generalization to unseen domains remains challenging. Recent studies have introduced domain generalization strategies and alternative-view synthesis techniques. Wang et al.~\cite{r57} attempt to construct a domain-agnostic shared latent space to mitigate distribution gaps across domains. However, they remain sensitive to appearance changes and severe perspective distortion. More importantly, the limited ability of 2D imagery to preserve stable geometric and structural cues restricts its generalization to unfamiliar scenes.

To address the above limitations, we propose \textbf{GeoLink}, a 3D-aware semantic-consistent framework for generalizable CVGL. Specifically, GeoLink first offline reconstructs scene point clouds from multi-view drone images and encodes them into discriminative 3D structural features. These 3D representations are then used as view-consistent anchors to guide 2D representation learning in two complementary ways. First, a \emph{Geometric-aware Semantic Refinement} module suppresses structurally redundant and viewpoint-biased information in 2D features under 3D guidance. Second, a \emph{Unified View Relation Distillation} module transfers instance-level relational knowledge in 3D space to 2D image features, enhancing inference efficiency without introducing additional computation. In this way, GeoLink improves robustness to both severe cross-view inconsistency and domain shift.

Based on the above analysis, the main contributions of this work are summarized as follows:
\begin{itemize}
    \item We propose \textbf{GeoLink}, a 3D-aware framework for generalizable CVGL, which introduces explicit scene-level 3D structural priors as a stable cross-view anchor under severe viewpoint variation and domain shift.
    
    \item We employ \textbf{Geometric-aware Semantic Refinement} module to suppress nuisance structural redundancy in 2D representations and \textbf{Unified View Relation Distillation} module to transfer view-consistent relational knowledge from 3D space to 2D features.
    
    \item GeoLink preserves a \textbf{2D-only inference pipeline}, while consistently improving cross-area and multi-weather generalization performance across CVGL benchmarks, demonstrating a favorable trade-off between effectiveness and practical deployment.
\end{itemize}

\section{Related Work}

\subsection{Cross-view Geo-localization}
Advances in deep learning have promoted dual-branch architectures that learn view-shared features for cross-view matching~\cite{r18,r20}. Transformer-based methods~\cite{r22} further improve contextual correspondence modeling through long-range feature interactions~\cite{r23,r76, r77}. Despite promising in-domain performance, these supervised methods typically rely on large-scale paired annotations, limiting their transferability to unseen domains. Recent generalizable CVGL methods have begun to address this limitation. For example, InfoGeo employs object-centric learning to capture transferable cross-view semantics~\cite{r74}. However, they primarily focus on 2D appearance and object-level representations, while largely overlooking scene-level geometry and cross-view structural relations that may provide more stable and transferable cues under domain shift. Specifically, we introduce 3D representations as stable structural anchors to regularize 2D cross-view features under severe viewpoint variation and domain shift.

\subsection{Multi-view Synthesis in CVGL}
Due to the significant viewpoint gap between query and gallery images, traditional CVGL methods face substantial challenges, making viewpoint transformation and synthesis promising solutions. BEV-CV~\cite{r27} narrows this gap by converting ground panoramas into semantic BEV representations. Other works~\cite{r28, r30} enrich viewpoint information and alleviate data scarcity using generative models such as GANs~\cite{r31} and diffusion models~\cite{r32}. Beyond direct synthesis, SkyLink~\cite{r75} uses UAV-mediated 3D scene representations to bridge street and satellite views through self-supervised and cross-view contrastive learning. Recent methods~\cite{r13} further employ 3D Gaussian Splatting to synthesize alternative viewpoints. However, these approaches mainly use 3D information for viewpoint bridging or generation, without explicitly leveraging scene geometry to regularize transferable 2D representations. Therefore, we instead use offline-reconstructed point clouds as training-only structural anchors while preserving 2D-only inference.

\section{The Proposed Method}
\begin{figure*}[ht!]
	\centering
	\includegraphics[width=0.975\textwidth]{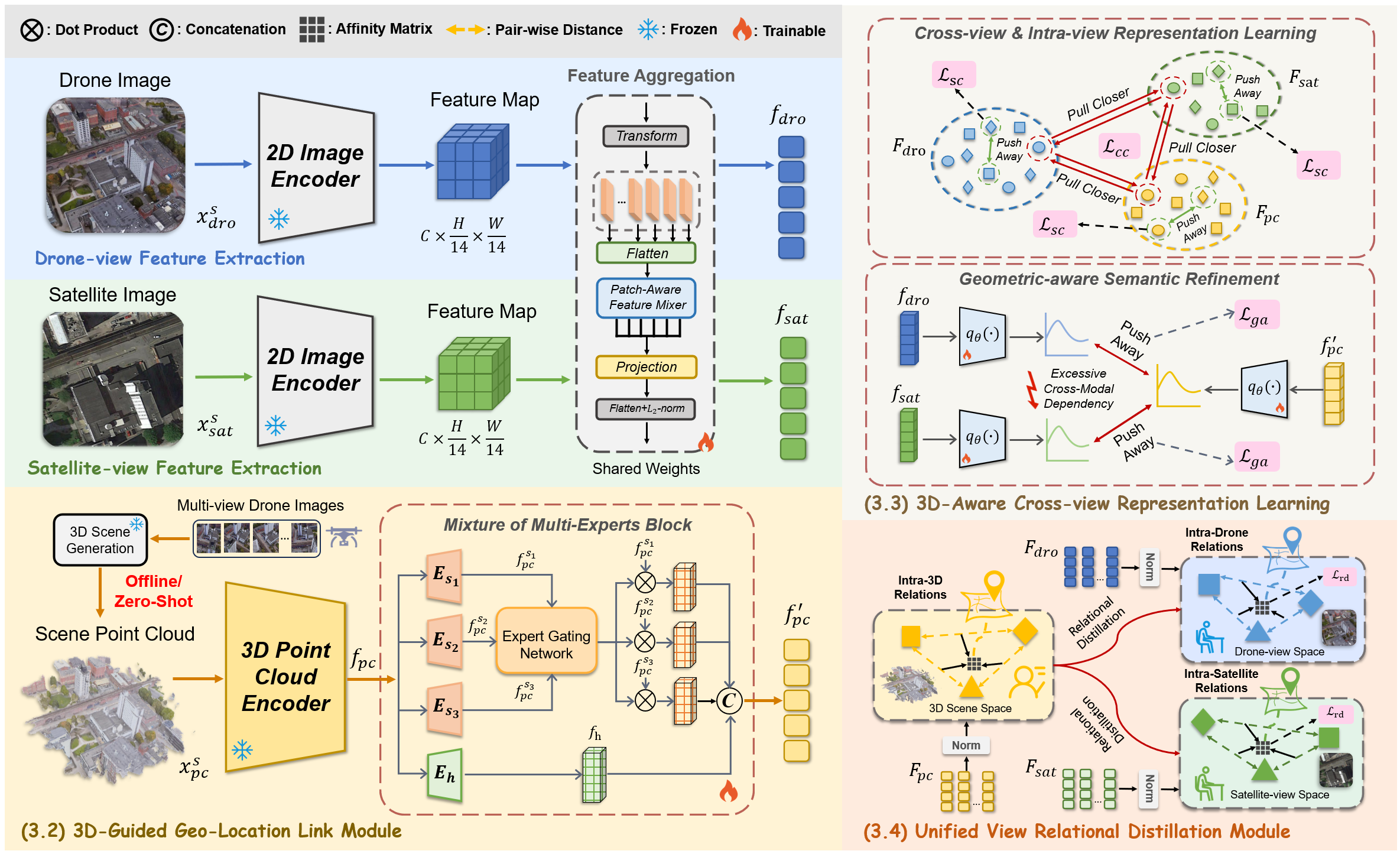}
	\caption{The overview of the proposed framework. The key idea of GeoLink is to use scene-level 3D structure as a stable structural anchor for generalizable CVGL. Multi-view drone images are first converted into offline scene point clouds, which provide view-consistent structural priors during training. These 3D anchors guide 2D retrieval learning at two complementary levels: they regularize 2D features to reduce potentially view-biased dependencies, and they transfer instance-level structural relations from 3D space to 2D image features.}
	\label{fig:overview}
\Description{Overview diagram of the GeoLink framework with 3D-guided linking, semantic refinement, and relation distillation.}
    \vspace{-1.em}
\end{figure*}

\subsection{Problem Definition and Overview}
Let $D_s$ denote the paired source-domain dataset, where drone-view images are used as queries and satellite-view images form the gallery. We denote the source-domain drone-view and satellite-view image sets by $X_s^{dro}$ and $X_s^{sat}$, respectively, where $X_s^{dro}=\{x_i^{dro}\in\mathbb{R}^{C\times H\times W}\}_{i=1}^{N^{dro}}$ and $X_s^{sat}=\{x_j^{sat}\in\mathbb{R}^{C\times H\times W}\}_{j=1}^{N^{sat}}$. The goal of generalizable CVGL is to learn robust cross-view representations from the source domain that remain transferable to an unseen target domain $D_t$ with image sets $X_t^{dro}$ and $X_t^{sat}$.

Figure~\ref{fig:overview} illustrates the overall framework of GeoLink. The proposed method consists of three key components: 3D-Guided Geo-Location Link Module, 3D-aware cross-view representation learning, and Unified View Relation Distillation Module. Specifically, we build upon the baseline in~\cite{r48}, which adopts 2D image encoders with feature aggregation to extract 2D features $F^{dro} \in \mathbb{R}^{N^{dro} \times D}$ and $F^{sat} \in \mathbb{R}^{N^{sat} \times D}$ from $X_s^{dro}$ and $X_s^{sat}$, respectively, where $D$ is the feature dimension. Based on this pipeline, GeoLink further introduces explicit 3D structural guidance and cross-view semantic regularization to enhance the discriminability and transferability of the learned representations.

\subsection{3D-Guided Geo-Location Link Module}
The core challenge in CVGL lies in preserving semantic consistency under extreme viewpoint changes. Although geometry-based methods establish cross-view correspondences~\cite{r40, r41}, they fail when key landmarks are occluded. Leveraging the ability of modern 3D reconstruction techniques to generate point clouds from multi-view drone imagery in benchmarks, we propose a 3D-Guided Geo-Location Link Module that addresses these limitations by exploiting domain-invariant features through 3D geometric and structural guidance, as shown in Figure~\ref{fig:overview} (a). The framework begins with the frozen VGGT model for 3D scene generation. Given input drone images $X^{dro}_{s}$, the frozen VGGT model $F_{VGGT}(\cdot)$ predicts corresponding point clouds $X^{pc}_{s}$:
\begin{equation}
    X^{pc}_{s} = F_{VGGT}(X^{dro}_{s}),
\end{equation}
where 3D point clouds are denoted as $X^{pc}_{s}=\{x^{pc}_{s}\in(p_1, p_2,...,p_M)\}$, where $p_i=(x_i, y_i, z_i)_{i}^{M}$ with $M$ points.

\subsubsection{3D Point Cloud Encoder}
To prevent excessive resource consumption during training, we employ a non-parametric point cloud encoder as the 3D Point Cloud Encoder. We follow Point-NN~\cite{r37}, which stacks non-learnable operations (FPS, k-NN, and pooling) into a four-stage pyramid for progressive downsampling. The input at stage \(s\) is:
\begin{equation}
P_{s} = \left\{ (\mathrm{p}_i, \mathrm{f}_i) \mid \mathrm{p}_i \in \mathbb{R}^{1\times3}, \mathrm{f}_i \in \mathbb{R}^{1\times C} \right\}_{i=1}^{M}, 
\end{equation}
where \(\mathrm{f}_i\) is the feature of \(\mathrm{p}_i\) from the previous stage.Each stage applies FPS, k-NN grouping, and pooling to progressively update the features, followed by global pooling to obtain \(F_{pc}\).


\subsubsection{Mixture of Multi-Expert Block}
Although the 3D encoder can effectively capture global scene features, limited drone-view images often lead to sparse point clouds lacking fine-grained details. To address this, we expand the receptive field in the 3D view by introducing a Mixture of Multi-Expert (MME) module for adaptive multi-expert fusion. 

Inspired by the Mixture of Experts (MoE) paradigm~\cite{r46}, we introduce a Mixture of Multi-Expert (MME) block for adaptive 3D modeling. The motivation is that point clouds reconstructed from limited drone views are often sparse and cannot simultaneously preserve fine-grained local details and broader structural context.
 
Specifically, three experts, denoted as $E_{s_1}(\cdot)$, $E_{s_2}(\cdot)$, and $E_{s_3}(\cdot)$, are used to capture geometric patterns under a wider receptive field, enabling the extraction of complementary structural features. An expert gating network $G(\cdot)=\{g_1(\cdot), g_2(\cdot), g_3(\cdot)\}$ is introduced to adaptively aggregate these responses. Given the input features $F_{pc}\in\mathbb{R}^{B\times D}$, where $B$ is the batch size and $D$ is the feature dimension, the gating network predicts expert-wise weights in $\mathbb{R}^{B\times 3}$. The output of each expert is then reweighted as
\begin{equation}
F^{s_i}_{pc}=g_i(F_{pc})\cdot E_{s_i}(F_{pc}), \quad i=1,2,3.
\end{equation}

Besides, a shared expert $E_h(\cdot)$ with a \textit{$1\times1$} convolution is employed to refine the holistic representation and encourage information sharing across experts. It produces a holistic feature $F^h_{pc}$, which is concatenated with the outputs of the three experts to form the final 3D representation:
\begin{equation}
F_{pc}^{'}=\mathrm{Concat}(F^{s_1}_{pc}, F^{s_2}_{pc}, F^{s_3}_{pc}, E_h(F_{pc})).
\end{equation}

By combining adaptive scale selection with shared feature refinement, the proposed MME block captures both local geometric details and broader structural context, leading to more discriminative and transferable 3D representations for CVGL.

\subsection{3D-aware Cross-view Representation Learning}
Intuitively, drone images capture fine-grained local details, while satellite views provide coarse global context. Most CVGL methods align views via contrastive learning but often overlook view heterogeneity, where some perspectives contain unique information or exhibit semantic divergence. Additionally, 3D point clouds offer view-consistent geometric information across views. As shown in Figure~\ref{fig:overview} (b), we employ the 3D-aware Cross-view Representation Learning to tackle these limitations.

\subsubsection{Cross-view \& Intra-view Representation Learning}
To learn view-consistent discriminative features, we jointly employ cross-view contrastive learning and intra-view discriminative learning within a unified framework using the InfoNCE objective~\cite{r51}. 

Specifically, cross-view contrastive learning aims to bridge the representation gap across different views by pulling closer together queries and their corresponding samples from other views. The formulation is based on the InfoNCE loss:
\begin{equation}
\mathcal{L}(q, R)_{Info} = -\log \frac{\exp(q \cdot r_+ / \tau)}{\sum_{i=0}^{R} \exp(q \cdot r_i / \tau)},
\end{equation}
where $q$ denotes the query, $R$ represents a set of reference samples, and $r_+$ is the positive sample corresponding to $q$. For drone-satellite matching, the cross-view contrastive loss is defined as:
\begin{equation}
\small
\mathcal{L}_{cc} = \mathcal{L}(dro, sat)_{Info} + \mathcal{L}(dro, pc)_{Info} + \mathcal{L}(sat, pc)_{Info},
\end{equation}
where $dro$, $sat$, and $pc$ denote drone-view imagery, satellite imagery, and scene point clouds, respectively.

To enhance instance-level discrimination within each view, we introduce intra-view contrastive learning, treating each instance as a distinct anchor to capture fine-grained differences. For a given anchor, its feature is positive, while other instances in the same batch serve as negatives. The intra-view loss is calculated as:
\begin{equation}
\small
\begin{aligned}
\mathcal{L}_{sc} = \mathcal{L}(F_{dro}, F_{dro})_{Info} + \mathcal{L}(F_{sat}, F_{sat})_{Info} + \mathcal{L}(F_{pc}, F_{pc})_{Info}.
\end{aligned}
\end{equation}
In this way, $L_{sc}$ preserves instance-wise separability within each view, complementing cross-view alignment and preventing visually similar locations from being overly compressed.

By jointly optimizing \(L_{cc}\) and \(L_{sc}\), the model promotes cross-view semantic alignment while preserving instance-level discrimination. This complementary design improves more robust and fine-grained representations for cross-view matching.

\subsubsection{Geometry-aware Semantic Refinement.}
Cross-view contrastive learning is widely used to align paired representations, but may retain correlations associated with training viewpoints and scene layouts. Prior studies have shown that 3D priors can improve 2D representation learning~\cite{r1,r2}. However, without explicit dependency control, the 2D encoder may overfit non-transferable geometric patterns. We therefore use the reconstructed 3D representation as an information reference to regulate, rather than indiscriminately maximize, dependence on each 2D view.

Given the refined 3D feature $F_{pc}'$ and a 2D feature $F_v$, where $v\in\{\mathrm{dro},\mathrm{sat}\}$, we employ vCLUB~\cite{r56} to obtain a tractable variational estimate of their mutual information:
\begin{equation}
\begin{aligned}
\mathcal{L}_{\mathrm{vCLUB}}^{(v)}
={}&
\mathbb{E}_{p(F_{pc}',F_v)}
\left[
\log q_{\theta}^{(v)}(F_v\mid F_{pc}')
\right] \\
&-
\mathbb{E}_{p(F_{pc}')}
\mathbb{E}_{p(F_v)}
\left[
\log q_{\theta}^{(v)}(F_v\mid F_{pc}')
\right],
\end{aligned}
\label{eq:club}
\end{equation}
where $q_{\theta}^{(v)}(F_v\mid F_{pc}')$ approximates the conditional distribution of the 2D representation given its 3D structural reference. The first term evaluates paired cross-modal features, whereas the second is estimated from independently sampled features. The final geometry-aware regularization is defined as
\begin{equation}
\mathcal{L}_{ga}
=
\mathcal{L}_{\mathrm{vCLUB}}^{(\mathrm{dro})}
+
\mathcal{L}_{\mathrm{vCLUB}}^{(\mathrm{sat})}.
\label{eq:geometry_regularization}
\end{equation}

Importantly, $\mathcal{L}_{ga}$ is jointly optimized with cross-view contrastive learning and relational distillation. The former preserves location-discriminative agreement, while the latter transfers view-consistent 3D relations to the image space. Within this joint objective, $\mathcal{L}_{ga}$ limits excessive 2D--3D coupling without enforcing global semantic decoupling, thereby improving transferability to unseen domains.

\subsection{Unified View Relational Distillation Module}
Based on the above analysis, 3D representations provide view-consistent spatial understanding unattainable from 2D views alone. Inspired by previous distillation methods~\cite{r58}, we propose a Unified View Relation Distillation module that transfers cross-modal knowledge from 3D space to 2D features, as shown in Figure~\ref{fig:overview} (c). This guidance helps image encoders retain view-invariant spatial structure and enhances generalization across diverse scenarios, and avoids extra multi-modal usage during inference.

We construct pairwise affinity matrices for the teacher and student features based on normalized Euclidean distances $\|\cdot \|_{2}$. For the 3D teacher feature set $T=\{t_i\}_{i=1}^{B}$ and the 2D student feature set $S^{(v)}=\{s_i^{(v)}\}_{i=1}^{B}$ from view $v$, the normalized affinities $A_T$ and $A_S$ are defined as follows:
\begin{equation}
\left\{
\begin{aligned}
A_T(i,j)
&=
\frac{\|t_i-t_j\|_2}
{\sum_{m\neq n}\|t_m-t_n\|_2}, \\[3pt]
A_S^{(v)}(i,j)
&=
\frac{\|s_i^{(v)}-s_j^{(v)}\|_2}
{\sum_{m\neq n}\|s_m^{(v)}-s_n^{(v)}\|_2},
\end{aligned}
\right.
\quad i\neq j.
\label{eq:ed}
\end{equation}

In addition, the relation distillation loss $\mathcal{L}^{(v)}_{rd}$ for view $v$ is computed using Mean Square Error (MSE):
\begin{equation}
L_{rd}^{(v)}=
\frac{1}{B(B-1)}\sum_{i\neq j}
\left\|A_S^{(v)}(i,j)-A_T(i,j)\right\|_2^2.
\label{eq:urd-euc-loss-view}
\end{equation}

We obtain the final objective by aggregating the total relation distillation losses $\mathcal{L}_{d}$ from both views:
\begin{equation}
\mathcal{L}_{\text{rd}} = \mathcal{L}^{(sat)}_{\text{rd}}+\mathcal{L}^{(dro)}_{rd}.
\label{eq:urd-euc-final}
\end{equation}

The objective drives the students' views to acquire 3D Relation-aware representations, ensuring both semantic and spatial alignment in CVGL.

\subsection{Optimization and Inference}
Finally, the aforementioned loss functions of the cross-view and intra-view CL, Geometric-aware semantic refinement, and unified view relation distillation are integrated into a unified framework. The GeoLink is trained using the following loss function:
\begin{equation}
\mathcal{L}_{total} = \mathcal{L}_{cc}+\mathcal{L}_{rd}+\mathcal{L}_{ga}+\lambda \mathcal{L}_{sc},
\end{equation}
where $\lambda$ is a weighting factor that balances the contribution of intra-view contrastive learning (set to 4.0).

During inference, both the query and gallery images are only passed into the image encoder to extract features for retrieval.

\section{Experiments}

\begin{table}[t]
\caption{Statistics of the training CVGL benchmarks.}
\label{uav-dataset}
\centering
\footnotesize
\setlength{\tabcolsep}{6.5pt}
\renewcommand{\arraystretch}{1.0}
\begin{tabular}{@{}lccc@{}}
\toprule
\rowcolor{gray!15}
\textbf{Dataset} & \textbf{IDs} & \textbf{Images (D/S)} & \textbf{Altitude Range} \\
\midrule
University-1652 & 701 & 37,854 / 701 & 121m--256m \\
SUES-200 & 120 & 24,000 / 480 & 150m--300m \\
DenseUAV & 2,256 & 6,768 / 13,536 & 80m--100m \\
\bottomrule
\end{tabular}
\end{table}

\begin{table*}[t]
\caption{Cross-area comparison (\%) in the Drone$\rightarrow$Satellite setting. Best and second-best results are highlighted in \best{bold} and \second{underline}, respectively.}
\label{dg_d2s}
\centering
\scriptsize
\setlength{\tabcolsep}{3.5pt}
\renewcommand{\arraystretch}{1.1}
\resizebox{\textwidth}{!}{%
\begin{tabular}{@{}llcccccccccccc@{}}
\toprule
\multirow{2}{*}{\textbf{Method}} & \multirow{2}{*}{\textbf{Venue}} & \multicolumn{2}{c}{\textbf{U1652$\rightarrow$SUES}} & \multicolumn{2}{c}{\textbf{DUAV$\rightarrow$SUES}} & \multicolumn{2}{c}{\textbf{SUES$\rightarrow$U1652}} & \multicolumn{2}{c}{\textbf{DUAV$\rightarrow$U1652}} & \multicolumn{2}{c}{\textbf{U1652$\rightarrow$DUAV}} & \multicolumn{2}{c}{\textbf{SUES$\rightarrow$DUAV}} \\
\cmidrule(lr){3-4}\cmidrule(lr){5-6}\cmidrule(lr){7-8}\cmidrule(lr){9-10}\cmidrule(lr){11-12}\cmidrule(lr){13-14}
& & \textbf{R@1} & \textbf{AP} & \textbf{R@1} & \textbf{AP} & \textbf{R@1} & \textbf{AP} & \textbf{R@1} & \textbf{AP} & \textbf{R@1} & \textbf{AP} & \textbf{R@1} & \textbf{AP} \\
\midrule
MCCG & TCSVT'23 & 70.31 & 74.58 & 80.09 & 83.18 & 34.70 & 39.47 & 50.52 & 55.35 & 17.54 & 13.01 & 16.66 & 14.68 \\
Sample4Geo & ICCV'23 & 82.03 & 85.12 & 78.94 & 81.86 & 62.13 & \second{66.48} & 37.90 & 42.54 & 33.46 & 22.28 & 24.56 & 17.63 \\
DAC & TCSVT'24 & 87.65 & 89.80 & 86.44 & 88.70 & 61.52 & 65.62 & 45.41 & 50.03 & 37.81 & 26.99 & 25.37 & 18.02 \\
MFRGN & ACM MM'24 & 78.28 & 81.75 & 67.22 & 71.14 & 35.58 & 40.95 & 37.95 & 41.73 & 23.60 & 16.08 & 22.55 & 14.51 \\
CV-Cities & JSTARS'24 & 90.27 & 91.77 & \second{89.67} & \second{91.34} & 57.69 & 62.22 & 36.76 & 32.43 & \second{38.87} & \best{28.44} & 21.60 & 16.74 \\
QDFL & TGRS'25 & \second{92.01} & \second{92.82} & 87.33 & 89.10 & \second{63.08} & 66.36 & \second{65.83} & \second{71.50} & 31.40 & 20.34 & \second{27.60} & \second{19.38} \\
Game4Loc & AAAI'25 & 86.75 & 89.07 & 67.19 & 71.40 & 44.86 & 49.35 & 26.47 & 31.05 & 22.88 & 18.54 & 20.44 & 17.32 \\
MMGeo & ICCV'25 & 85.01 & 87.60 & 79.05 & 82.05 & 48.76 & 53.77 & 33.92 & 38.28 & 17.76 & 13.62 & 18.08 & 13.17 \\
\addlinespace[2pt]
\rowcolor{geolinkshade}
GeoLink & This work & \best{93.44} & \best{94.38} & \best{93.25} & \best{94.31} & \best{63.41} & \best{67.69} & \best{68.28} & \best{72.26} & \best{39.30} & \second{27.93} & \best{33.03} & \best{23.12} \\
\bottomrule
\end{tabular}%
}
\end{table*}

\subsection{Experiment Setup}

\textbf{Datasets and Evaluation Protocol.}
We evaluate on University-1652~\cite{r10}, SUES-200~\cite{r11}, and DenseUAV~\cite{r12}, denoted U1652, SUES, and DUAV. For offline VGGT reconstruction, all drone-view images are used to generate point clouds: about 50 images per scene for U1652 and SUES, and three height-varied UAV images for DUAV. SUES-200 results are averaged over four altitudes (150 m, 200 m, 250 m, and 300 m). We evaluate Recall@K (R@K), indicating whether the ground-truth match appears in the Top-$K$ results, and Average Precision (AP). We report Drone$\rightarrow$Satellite and Satellite$\rightarrow$Drone protocols, alternately using each view as query and gallery.

\textbf{Implementation Details.}
The experiments are conducted on a single NVIDIA 4090 GPU using the PyTorch framework. We use DINOv2-Base~\cite{r55} and Point-NN~\cite{r37} as the 2D and 3D backbones, respectively. And Mixer~\cite{r73} is used for feature aggregation. All images are resized to $448\times448$, while 2D features are aggregated into 4096-dimensional vectors. For point clouds, we sample 1024 points without colors. 3D features are downsampled to 4096 dimensions through the proposed MME block. The hidden dimension of $q_{\theta}(\cdot)$ in vCLUB is set to 4096. The model is trained for $40$ epochs with a batch size of $8$. AdamW~\cite{r72} is adopted as optimizer with an initial learning rate of $6 \times 10^{-4}$, which follows a cosine decay schedule with a 0.1-epoch warmup and gradually decays to $1 \times 10^{-4}$. The weighting factor $\lambda$ is empirically set to 4.0.

\subsection{Comparison with SOTA methods}
To evaluate GeoLink, we compare it with recent state-of-the-art methods in the cross-area setting, including MCCG~\cite{r61}, Sample4Geo~\cite{r9}, DAC~\cite{r19}, MFRGN~\cite{r62}, QDFL~\cite{r64}, Game4Loc~\cite{r70}, MMGeo~\cite{r71}, and CV-Cities~\cite{r48}. Table~\ref{dg_d2s} shows that GeoLink achieves the best overall performance under the \textit{Drone}$\rightarrow$\textit{Satellite} protocol, with greater advantages in challenging transfer settings. On DUAV$\rightarrow$U1652, GeoLink improves R@1 from 65.83\% to 68.28\% and AP from 71.50\% to 72.26\%, indicating that sparse point clouds reconstructed from limited UAV views still provide effective structural priors. On SUES$\rightarrow$DUAV, it raises R@1 from 27.60\% to 33.03\% and AP from 19.38\% to 23.12\%, demonstrating robust generalization across substantial domain shifts.

Table~\ref{dg_s2d} reports the reverse \textit{Satellite}$\rightarrow$\textit{Drone} protocol, where GeoLink again achieves the strongest overall performance. On U1652$\rightarrow$SUES, it improves R@1 and AP by 1.57\% and 4.43\%, respectively. On DUAV$\rightarrow$SUES, despite an R@1 0.28\% below CV-Cities, GeoLink obtains the best AP with a 0.91\% gain. Overall, these results demonstrate that the proposed 3D-aware framework enhances cross-view alignment and instance-level discrimination under diverse cross-domain shifts.

\begin{table}[t]
\caption{Cross-area comparison (\%) in the Satellite$\rightarrow$Drone setting. Best and second-best results are highlighted in \best{bold} and \second{underline}, respectively.}
\label{dg_s2d}
\centering
\scriptsize
\setlength{\tabcolsep}{3.5pt}
\renewcommand{\arraystretch}{1.1}
\resizebox{\columnwidth}{!}{%
\begin{tabular}{@{}lcccccccc@{}}
\toprule
\multirow{2}{*}{\textbf{Method}} & \multicolumn{2}{c}{\textbf{U1652$\rightarrow$SUES}} & \multicolumn{2}{c}{\textbf{DUAV$\rightarrow$SUES}} & \multicolumn{2}{c}{\textbf{SUES$\rightarrow$U1652}} & \multicolumn{2}{c}{\textbf{DUAV$\rightarrow$U1652}} \\
\cmidrule(lr){2-3}\cmidrule(lr){4-5}\cmidrule(lr){6-7}\cmidrule(lr){8-9}
& \textbf{R@1} & \textbf{AP} & \textbf{R@1} & \textbf{AP} & \textbf{R@1} & \textbf{AP} & \textbf{R@1} & \textbf{AP} \\
\midrule
MCCG & 78.13 & 68.94 & 93.13 & 81.04 & 66.87 & 39.41 & 81.83 & 47.86 \\
Sample4Geo & 90.63 & 84.25 & 91.56 & 71.43 & 82.56 & 61.51 & 81.74 & 45.21 \\
DAC & 93.75 & 88.92 & \second{98.13} & \second{89.50} & 82.70 & 60.20 & 86.02 & 44.90 \\
MFRGN & 84.06 & 63.17 & 96.25 & 60.23 & 74.70 & 39.23 & \second{86.59} & 25.03 \\
CV-Cities & \second{96.56} & 88.07 & \best{98.41} & 88.13 & 78.31 & 56.85 & 83.58 & 71.66 \\
QDFL & 82.31 & 88.98 & 96.56 & 86.62 & \second{83.63} & \second{62.20} & 62.01 & \second{71.97} \\
Game4Loc & 93.44 & \second{89.22} & 88.44 & 67.40 & 77.64 & 45.81 & 71.75 & 26.19 \\
MMGeo & 91.25 & 84.61 & 93.13 & 75.76 & 76.53 & 47.44 & 72.33 & 35.43 \\
\addlinespace[2pt]
\rowcolor{geolinkshade}
GeoLink & \best{98.13} & \best{93.65} & \second{98.13} & \best{90.41} & \best{85.98} & \best{65.13} & \best{95.00} & \best{78.84} \\
\bottomrule
\end{tabular}%
}
\vspace{-10pt}
\end{table}

\subsection{Robustness Evaluation}
We evaluated robustness under degradations caused by diverse weather conditions, including fog-rain, fog-snow, rain-snow, and wind. As shown in Table~\ref{dg_weather}, the proposed model maintains robust results in University-1652$\to$SUES-200. Specifically, GeoLink consistently achieves the best or highly competitive performance under all adverse weather conditions. Compared with the baseline CV-cities, GeoLink delivers consistent improvements across all four weather conditions, with average gains of 6.67\% and 5.53\% in R@1 and AP under \textbf{fog-rain}, and 6.01\%/5.02\% under fog-snow, showing particularly clear advantages when the environmental degradations are more severe. Under \textbf{rain-snow}, GeoLink still maintains a noticeable margin of 3.79\% in R@1 and 2.86\% in AP, demonstrating stable robustness under coupled corruptions. Even under the \textbf{wind} setting, where the baseline is already strong, GeoLink achieves the best overall performance. These results indicate that 3D-aware training improves robustness to weather-induced appearance degradation and learns more transferable cross-view features.

\begin{table}[t]
\caption{Robustness evaluations on diverse weather cross-view transfer tasks (University-1652$\to$SUES-200, Drone$\to$Satellite). Best and second-best results are highlighted in \best{bold} and \second{underline}, respectively.}
\label{dg_weather}
\centering
\scriptsize
\setlength{\tabcolsep}{3.5pt}
\renewcommand{\arraystretch}{1.}
\resizebox{\columnwidth}{!}{%
\begin{tabular}{lcccccccc}
\toprule
\multirow{2}{*}{\textbf{Method}} & \multicolumn{2}{c}{\textbf{Fog-Rain}} & \multicolumn{2}{c}{\textbf{Fog-Snow}} & \multicolumn{2}{c}{\textbf{Rain-Snow}} & \multicolumn{2}{c}{\textbf{Wind}} \\ \cmidrule(lr){2-3}\cmidrule(lr){4-5}\cmidrule(lr){6-7}\cmidrule(lr){8-9} 
                                 & \textbf{R@1}      & \textbf{AP}       & \textbf{R@1}      & \textbf{AP}       & \textbf{R@1}       & \textbf{AP}       & \textbf{R@1}    & \textbf{AP}     \\ \midrule
Sample4Geo                       & 37.40             & 44.13             & 26.68             & 33.05             & 45.81              & 51.58             & 59.35           & 64.58           \\
DAC                              & 73.23             & 76.72             & 45.03             & 50.57             & 68.97              & 73.07             & 86.23           & 88.39           \\
CV-Cities                        & \second{75.11}             & \second{79.06}             & \second{72.33}             & \second{76.58}             & \second{85.24}              & \second{87.79}             & \second{88.11}           & \second{90.23}           \\
\addlinespace[2pt]
\rowcolor{geolinkshade}
GeoLink                          & \best{81.78}    & \best{84.59}    & \best{78.34}    & \best{81.53}    & \best{89.03}     & \best{90.70}    & \best{88.50}  & \best{90.43}  \\ \bottomrule
\end{tabular}}
\vspace{-10pt}
\end{table}

\subsection{Ablation Study}

\begin{table}[t]
\caption{Ablation study of the major components in GeoLink under two settings: SUES$\rightarrow$DUAV (Drone$\rightarrow$Satellite) and U1652$\rightarrow$SUES (Satellite$\rightarrow$Drone). $^{*}$ denotes the method trained with the same parameters as GeoLink.}
\label{ablation}
\centering
\small
\setlength{\tabcolsep}{8.0pt}
\renewcommand{\arraystretch}{1.}
\begin{tabular}{@{}c c c c c c c@{}}
\toprule
\multicolumn{3}{l}{\multirow{2}{*}{\textbf{Methods}}} & \multicolumn{2}{c}{\textbf{SUES$\rightarrow$DUAV}} & \multicolumn{2}{c}{\textbf{U1652$\rightarrow$SUES}} \\
\cmidrule(lr){4-5}\cmidrule(lr){6-7}
\multicolumn{3}{c}{} & \textbf{R@1} & \textbf{AP} & \textbf{R@1} & \textbf{AP} \\
\midrule
\multicolumn{3}{l}{Baseline$^{*}$} & 15.56 & 11.89 & 62.78 & 60.19 \\
\multicolumn{3}{l}{Baseline$^{*}$+MME} & 17.86 & 13.95 & 73.55 & 71.21 \\
\midrule
\textbf{$\mathcal{L}_{ga}$} & \textbf{$\mathcal{L}_{sc}$} & \textbf{$\mathcal{L}_{rd}$} & \textbf{R@1} & \textbf{AP} & \textbf{R@1} & \textbf{AP} \\
\midrule
\cmark & \xmark & \xmark & 21.80 & 14.72 & 75.39 & 72.56 \\
\xmark & \cmark & \xmark & 26.68 & 19.25 & 92.97 & 88.62 \\
\xmark & \xmark & \cmark & 22.61 & 15.60 & 90.81 & 85.97 \\
\cmark & \cmark & \xmark & 28.06 & 20.59 & 93.46 & 89.25 \\
\addlinespace[2pt]
\rowcolor{geolinkshade}
\cmark & \cmark & \cmark & \best{33.03} & \best{23.12} & \best{98.13} & \best{93.65} \\
\bottomrule
\end{tabular}
\vspace{-10pt}
\end{table}

We conduct ablation experiments to examine the contribution of each major component in GeoLink from two aspects: the effectiveness of the proposed MME block, and the role of each objective in the 3D-aware cross-view learning.

\subsubsection{Effect of the MME block.}
As shown in Table~\ref{ablation}, the proposed MME block consistently improves the baseline in both transfer settings. On the SUES$\rightarrow$DUAV, GeoLink achieves $15.56\%$ R@1 and $11.89\%$ AP. After introducing the proposed MME block, the performance increases to $17.86\%$ R@1 and $13.95\%$ AP, corresponding to gains of $2.30\%$ and $2.06\%$, respectively. Similar improvements are also observed on U1652$\rightarrow$SUES. These results indicate that adaptive Multi-Expert modeling in 3D space provides complementary structural cues that are helpful for handling cross-view variation and improving generalization.

\subsubsection{Effect of the proposed objectives.}
We further evaluate the contribution of each loss term in 3D-aware cross-view representation learning. Using only $L_{ga}$ improves the baseline, showing that suppressing structurally redundant dependencies under 3D guidance helps reduce view bias. Using only $L_{sc}$ yields larger gains, highlighting the importance of intra-view discrimination for feature compactness and instance separability. The relation distillation loss $L_{rd}$ also provides substantial improvements, confirming the effectiveness of transferring instance-level structural relations from 3D space to 2D features.

\begin{figure}[ht!]
	\centering
	\includegraphics[width=0.485\textwidth]{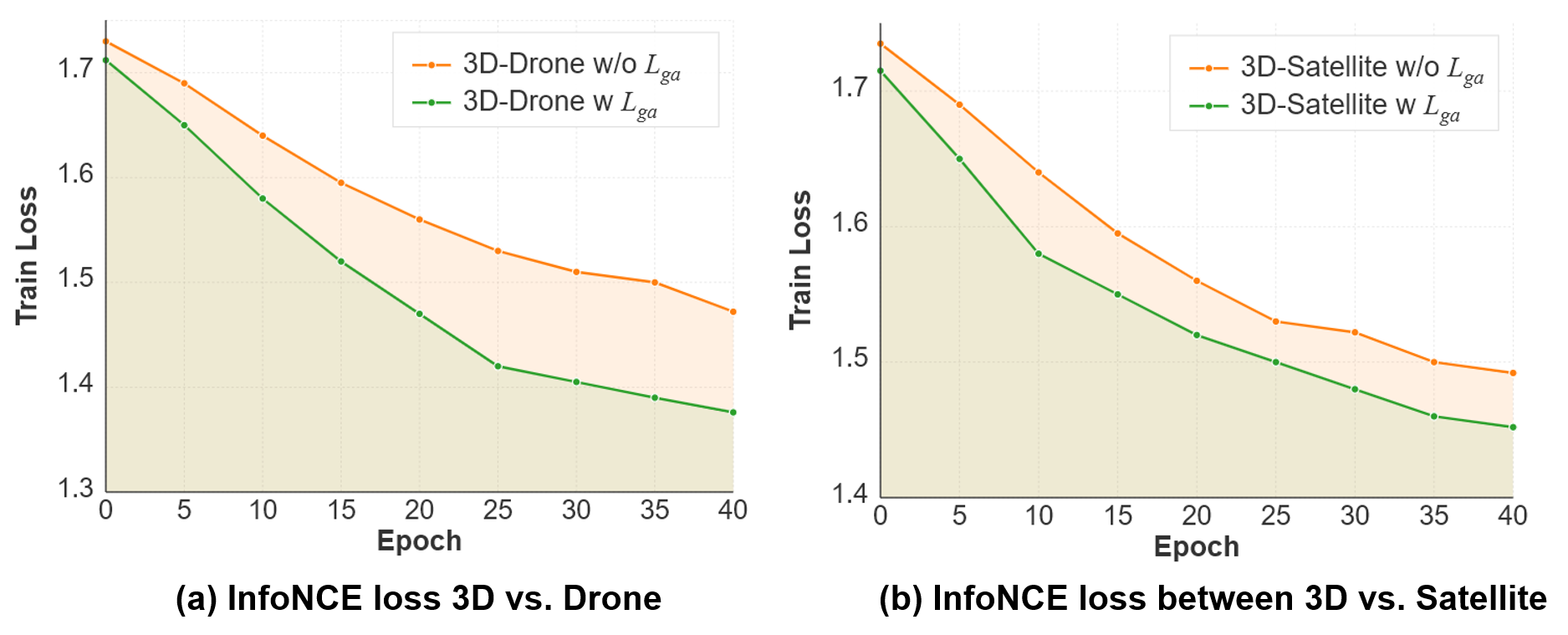}
	\caption{The impact of $\mathcal{L}_{ga}$ on the InfoNCE losses between 3D and Drone views, 3D and Satellite views, respectively.}
	\label{fig:vs}
\Description{Line charts of InfoNCE loss curves for 3D-drone and 3D-satellite view pairs.}
\end{figure}

Combining all three objectives achieves the best performance, with $33.03\%$ R@1 and $23.12\%$ AP on SUES$\rightarrow$DUAV, and $98.13\%$ R@1 and $93.65\%$ AP on U1652$\rightarrow$SUES. These results demonstrate their complementarity: the MME block strengthens 3D representations, $L_{ga}$ suppresses nuisance dependencies, $L_{sc}$ improves intra-view discrimination, and $L_{rd}$ transfers 3D relational knowledge. Moreover, the InfoNCE curves in Figure~\ref{fig:vs} decrease more steadily after geometric-aware refinement, indicating more stable and view-consistent learning under 3D structural regularization.

\subsection{Other Analysis}
\subsubsection{Computational Analysis.}
We compare representative methods in terms of trainable parameters and FLOPs to evaluate practical efficiency on a single NVIDIA 4090. As shown in Table~\ref{tab:computational_cost}, GeoLink has 90.50M trainable parameters and 91.72G FLOPs, matching CV-Cities in parameter size while introducing only marginally higher computation (91.72G vs. 91.67G). Although GeoLink requires slightly more resources than Sample4Geo and DAC, its overall complexity remains comparable. These results show that GeoLink achieves substantial performance gains without notable additional complexity, offering a favorable accuracy-efficiency trade-off.
\begin{table}[tbp]
\centering
\small
\caption{Comparisons of computational costs (Parameters and GFLOPs) of different methods.}
\label{tab:computational_cost}
\setlength{\tabcolsep}{1.0mm}
\begin{tabular}{l|ccc|cc}
\toprule
\rowcolor{gray!15}
 & \textbf{Sample4Geo} & \textbf{DAC} & \textbf{CV-Cities} & \textbf{GeoLink} \\
\midrule
Params (M) & 87.51 & 87.51 & 90.50 & 90.50\\
FLOPs (G)  & 72.60 & 80.22 & 91.67 &  91.72\\
\bottomrule
\end{tabular}
\vspace{-6pt}
\end{table}

\subsubsection{The Quality of 3D Reconstruction.}
To verify the quality of 3D point clouds, we first evaluate the intrinsic reconstruction quality by projecting them into BEV maps in Table~\ref{tab:vggt_3d_eval_results}. Since 3D GTs are unavailable, we compare BEV maps with satellite images using DINOv2 feature cosine similarity and Qwen3.5-VL pairwise prompts for semantic judgment (determine whether they belong to \textit{the same location} and share \textit{consistent spatial layouts}). These results show that they preserve reliable scene geometry and spatial cues.
\begin{table}[t]
\centering
\caption{Quality evaluation of VGGT-based Point Clouds.}
\vspace{-6pt}
\label{tab:vggt_3d_eval_results}
\small
\setlength{\tabcolsep}{6.0pt}
\renewcommand{\arraystretch}{0.95}
\begin{tabular}{cccc}
\toprule
\textbf{Metric} & \textbf{University-1652} & \textbf{SUES-200} & \textbf{DenseUAV} \\
\midrule
\textit{BEV Similarity} & $0.69 \pm 0.07$ & $0.78 \pm 0.05$ & $0.66 \pm 0.06$ \\
\textit{VLM Judgement}  &        $99.29\%$       &  $99.58\%$              & $97.78\%$ \\
\bottomrule
\end{tabular}
\vspace{-6pt}
\end{table}

\subsection{Qualitative Comparison}
For qualitative comparison, we visualize the PCA maps of feature maps and project the feature representations extracted by the CV-Cities and our method using t-SNE~\cite{r66}. 

\begin{figure}[ht!]
	\centering
	\includegraphics[width=0.485\textwidth]{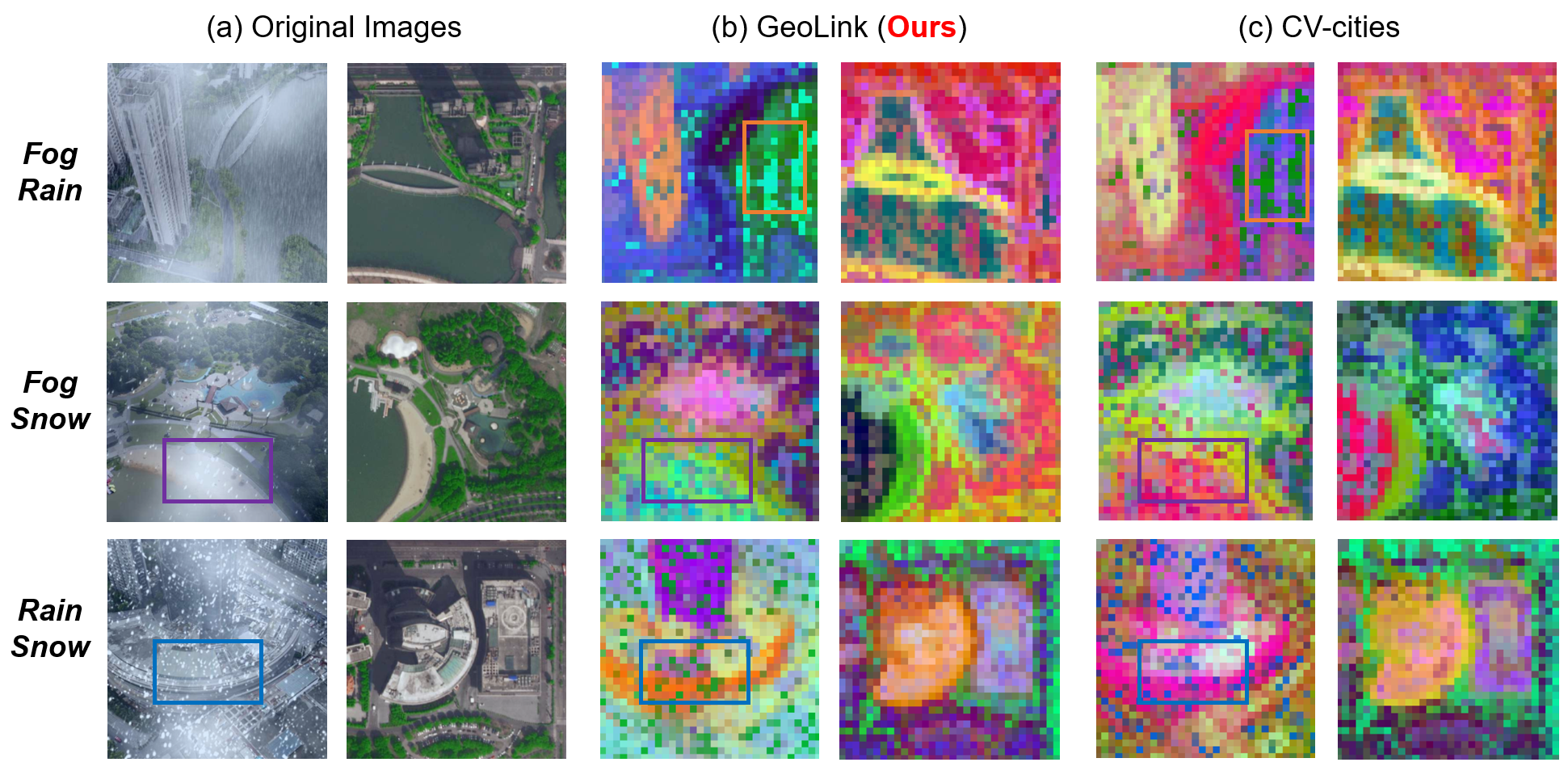}
	\caption{The comparisons of PCA visualizations on robustness evaluation (University-1652$\to$SUES-200@150m). The discriminative objects are marked by rectangular boxes.}
	\label{fig:pca}
\Description{PCA feature visualization comparing CV-Cities and GeoLink under adverse weather conditions.}
\end{figure}

As illustrated in Figure~\ref{fig:pca}, we visualize PCA maps from University-1652$\to$SUES-200@150m (Drone$\to$Satellite) under multiple weather conditions. Compared with CV-Cities, GeoLink produces more robust and discriminative responses under weather degradation. As highlighted by the colored boxes, GeoLink preserves clearer structural patterns and more concentrated activations around semantically meaningful regions (e.g., seashores and building blocks), despite severe fog-rain, fog-snow, and rain-snow corruption. In contrast, CV-Cities exhibits more scattered responses and is easily distracted by weather-induced noise, resulting in less distinctive cross-view correspondences. These results indicate that GeoLink better suppresses irrelevant interference and captures view-consistent structural cues, yielding more robust representations for localization in adverse environments.

\begin{figure}[t!]
	\centering
	\includegraphics[width=0.47\textwidth]{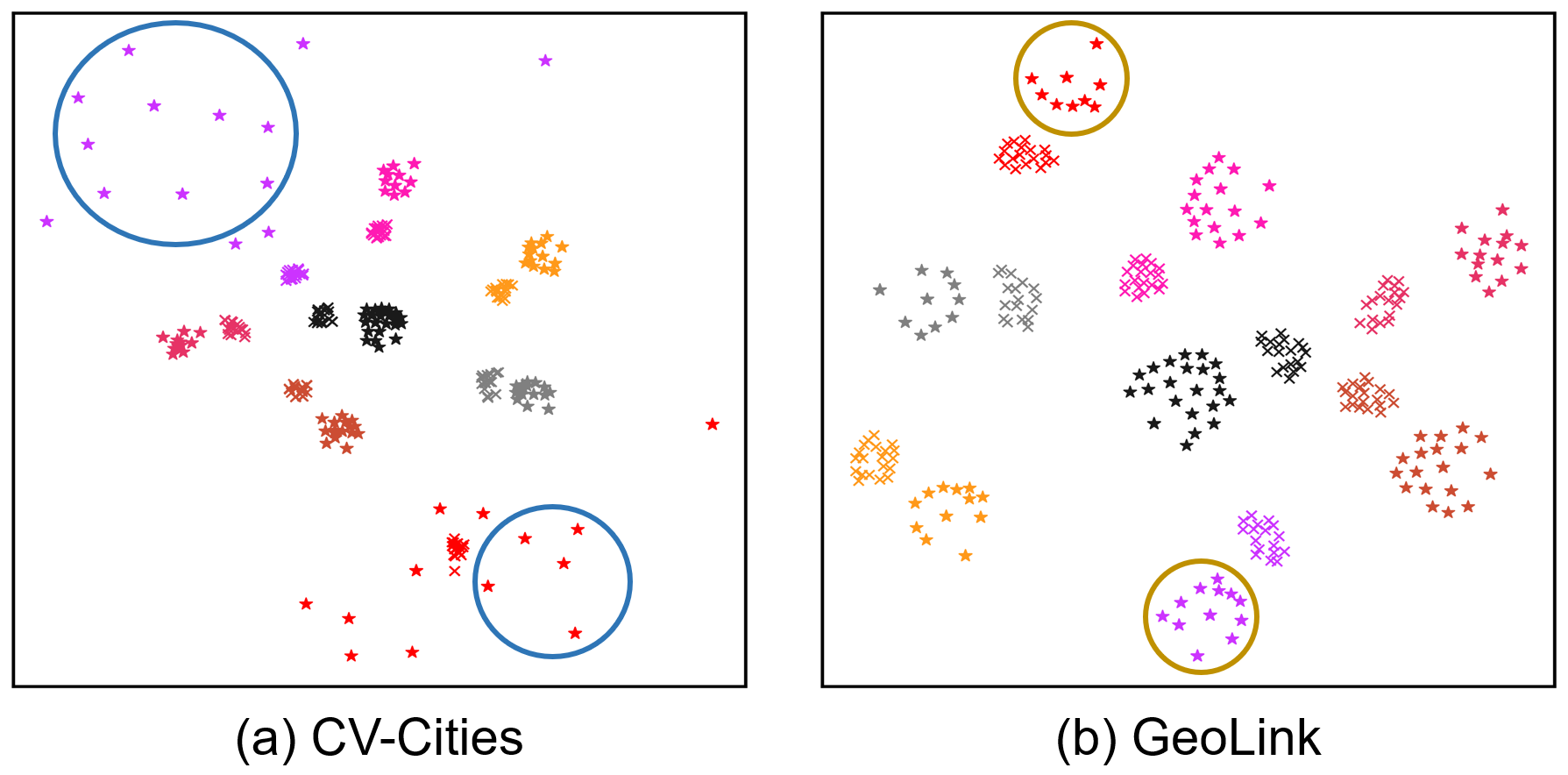}
	\caption{The T-SNE comparisons on CV-Cities and GeoLink across eight locations (each color represents a location). Stars and crosses denote drone and satellite views, respectively. Blue circles represent false samples.}
	\label{fig:7}
\Description{t-SNE visualization of drone and satellite feature embeddings for eight locations.}
\end{figure}

In addition, Figure~\ref{fig:7} highlights the superiority of GeoLink's representations over CV-Cities, attributed to its enhanced cross-view generalization. This is evident in the tighter clustering of same-class samples and improved alignment between drone and satellite perspectives. By bridging the viewpoint gap and preserving semantic consistency, GeoLink captures robust and discriminative features, demonstrating its effectiveness in CVGL tasks.

\subsection{Sensitivity Analysis}
\begin{table}[t]
\centering
\caption{Sensitivity Analysis of the number of experts in the MME block under two settings: SUES-200$\to$DenseUAV (Drone$\rightarrow$Satellite), SUES-200$\to$University (Satellite$\rightarrow$Drone). The best results are highlighted in bold.}
\small
\label{tab:expert_number}
\setlength{\tabcolsep}{5.85pt}
\begin{tabular}{c|cc|cc}
\toprule
\multirow{2}{*}{Experts} & \multicolumn{2}{c|}{SUES-200$\to$DenseUAV} & \multicolumn{2}{c}{SUES-200$\to$University} \\
\cmidrule(lr){2-3} \cmidrule(lr){4-5}
& R@1 & AP & R@1 & AP \\
\midrule
1  & 31.99 & 21.83 & 82.04 & 59.60 \\
\rowcolor{gray!15}
\textbf{3}  & \textbf{33.08} & \textbf{23.12} & \textbf{85.98} & \textbf{65.13} \\
7  & 28.19 & 20.34 & 80.25 & 55.77 \\
15 & 27.16 & 19.57 & 79.67 & 55.25 \\
\bottomrule
\end{tabular}
\end{table}

\begin{figure}[t!]
	\centering
	\includegraphics[width=0.485\textwidth]{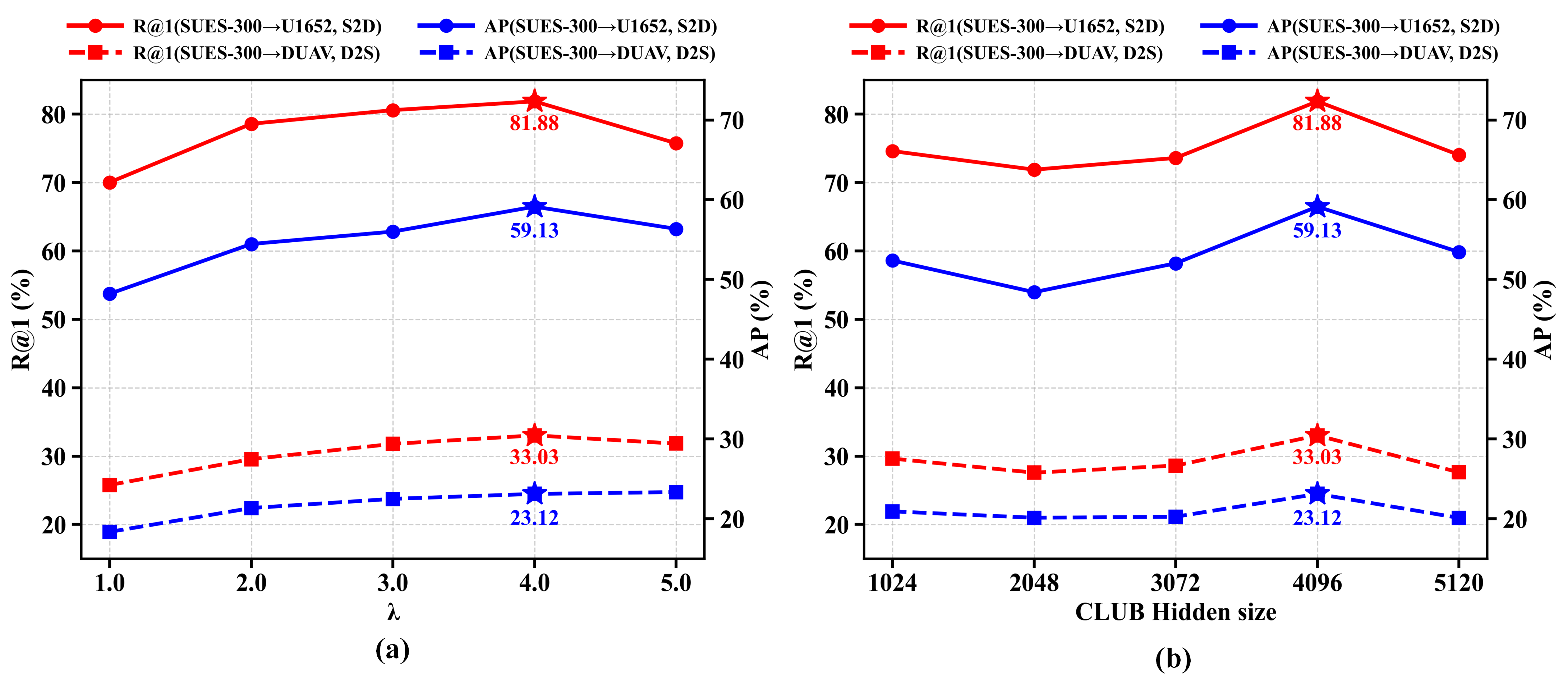}
	\caption{Sensitivity analysis of (a) the trade-off hyper-parameter $\lambda$ and (b) the hidden size of $q_{\theta}(\cdot)$ in vCLUB.}
	\label{fig:8}
\Description{Sensitivity plots for lambda and hidden size hyperparameters.}
\end{figure}

We further analyze the sensitivity of the number of experts in the MME block under two cross-domain settings. As shown in Table~\ref{tab:expert_number}, setting the expert number to 3 achieves the best performance on both SUES-200$\to$DenseUAV and SUES-200$\to$University, yielding 33.08\% R@1 / 23.12\% AP and 85.98\% R@1 / 65.13\% AP, respectively. When the number of experts is reduced to 1, the model cannot fully exploit geometric cues, leading to inferior performance. In contrast, using more experts (e.g., 7 or 15) causes clear performance degradation, suggesting that excessive experts may introduce redundant information and make feature aggregation less effective. These results indicate that a moderate number of experts provides the best trade-off between 3D representation capacity and discriminative feature learning.

In Figure~\ref{fig:8}, we present the sensitivity analysis of hyperparameters in two generalization scenarios. Figure~\ref{fig:8} (a) shows the sensitivity analysis of $\lambda$, which controls the weight of intra-view contrastive loss for semantic refinement. The model achieves peak performance at $\lambda = 4.0$, indicating an optimal trade-off between 3D-aware representation learning and global feature alignment. In Figure~\ref{fig:8} (b), increasing the hidden size of $q_{\theta}(\cdot)$ in vCLUB improves performance, saturating at 4096, suggesting that a more expressive estimator better captures cross-view mutual information. These results highlight GeoLink's robustness and strong generalization in cross-view scenarios.

\section{Conclusion}
In this paper, we propose \textbf{GeoLink}, a 3D-aware framework for generalizable CVGL. Different from prior methods that mainly rely on 2D appearance alignment, GeoLink introduces scene-level 3D structural priors as stable anchors for cross-view representation learning. Based on this design, the proposed MME block enhances 3D structural encoding, the Geometric-aware Semantic Refinement module suppresses structurally redundant and view-biased dependencies in 2D features, and the Unified View Relation Distillation module transfers 3D relational knowledge back to the image space. As a result, GeoLink learns more transferable and discriminative cross-view representations while preserving a 2D-only inference pipeline. Extensive experiments on zero-shot cross-domain and multi-weather settings demonstrate the effectiveness and robustness of the proposed method.

\begin{acks}
This work is supported by the Fundamental Research Funds for the Central Universities (Tongji University) under Grant No. 22120260380.
\end{acks}



\bibliographystyle{ACM-Reference-Format}
\bibliography{sample-base}

\end{document}